%% file: main.tex
\documentclass[11pt]{article}

\usepackage[a4paper, margin=2.5cm]{geometry}
\usepackage[T1]{fontenc}
\usepackage{lmodern}
\usepackage{setspace}
\usepackage[switch]{lineno}     %

\usepackage{amsmath, amssymb}
\usepackage{graphicx}
\usepackage{pdfpages}
\usepackage{booktabs}           %
\usepackage{caption}
\usepackage{newunicodechar}         %
\newunicodechar{–}{\textendash} \DeclareCaptionLabelSeparator{pipe}{ $|$ }   

\usepackage{authblk}

\usepackage[style=nature, backend=biber]{biblatex}
   \DeclareSourcemap{ \maps[datatype=bibtex]{ \map{\step[typesource=techreport, typetarget=report]} \map{\pernottype{dataset}\step[fieldset=urldate, null]} \map{\step[fieldset=eventtitle, null]} } }

\usepackage[hidelinks]{hyperref}

\title{\textbf{Prompt changes that raise individual-level agreement in language model survey simulations can worsen distributional fit}}

\author[1]{Yi-Wei Huang} \author[2,*]{Hsuan-Wei Lee} \affil[1]{Department of Information Management, National Taiwan University, Taipei, Taiwan} \affil[2]{Department of Biostatistics and Health Data Science, Lehigh University, Bethlehem, PA 18015, USA} \affil[*]{Corresponding author: hsl324@lehigh.edu} \date{}

\begin{document}

\maketitle

\begin{abstract}
\noindent Large language models can simulate survey responses, but a prompt change may affect individual predictions and aggregate distributions differently.
Three models predicted two vaccine attitudes of 340 respondents in the 2022 General Social Survey from six sociodemographic variables.
Six prompts varied where party identification appeared and how profiles were presented.
Under the base prompt, weighted agreement was below a cross-validated ordered logit benchmark in four of six model--item combinations.
Placing party identification first in a formal survey-style prompt raised weighted agreement on confidence in federal oversight in two models after false discovery rate correction.
The third model's smaller gain did not survive correction and came without lower squared error.
Averaged across party positions, formal prompts raised Claude Sonnet 4.6's weighted agreement but worsened its distributional fit for that attitude, unlike in the other models.
Consistent individual-level effects can coexist with model-specific distributional consequences, so each level needs separate validation.
\end{abstract}

\newpage

\input{sections/introduction}

\section*{Results}
\input{sections/results}

\section*{Discussion}
\input{sections/discussion}

\input{sections/methods}

\printbibliography

\section*{Acknowledgements}
This research received no external funding.

\section*{Author contributions}
Yi-Wei Huang: Methodology, Software, Formal analysis, Investigation, Visualization, Writing--original draft, Writing--review and editing.

\noindent Hsuan-Wei Lee: Conceptualization, Supervision, Project administration, Writing--review and editing.

\section*{Competing interests}
The authors declare no competing interests.

\clearpage
\includepdf[pages=-]{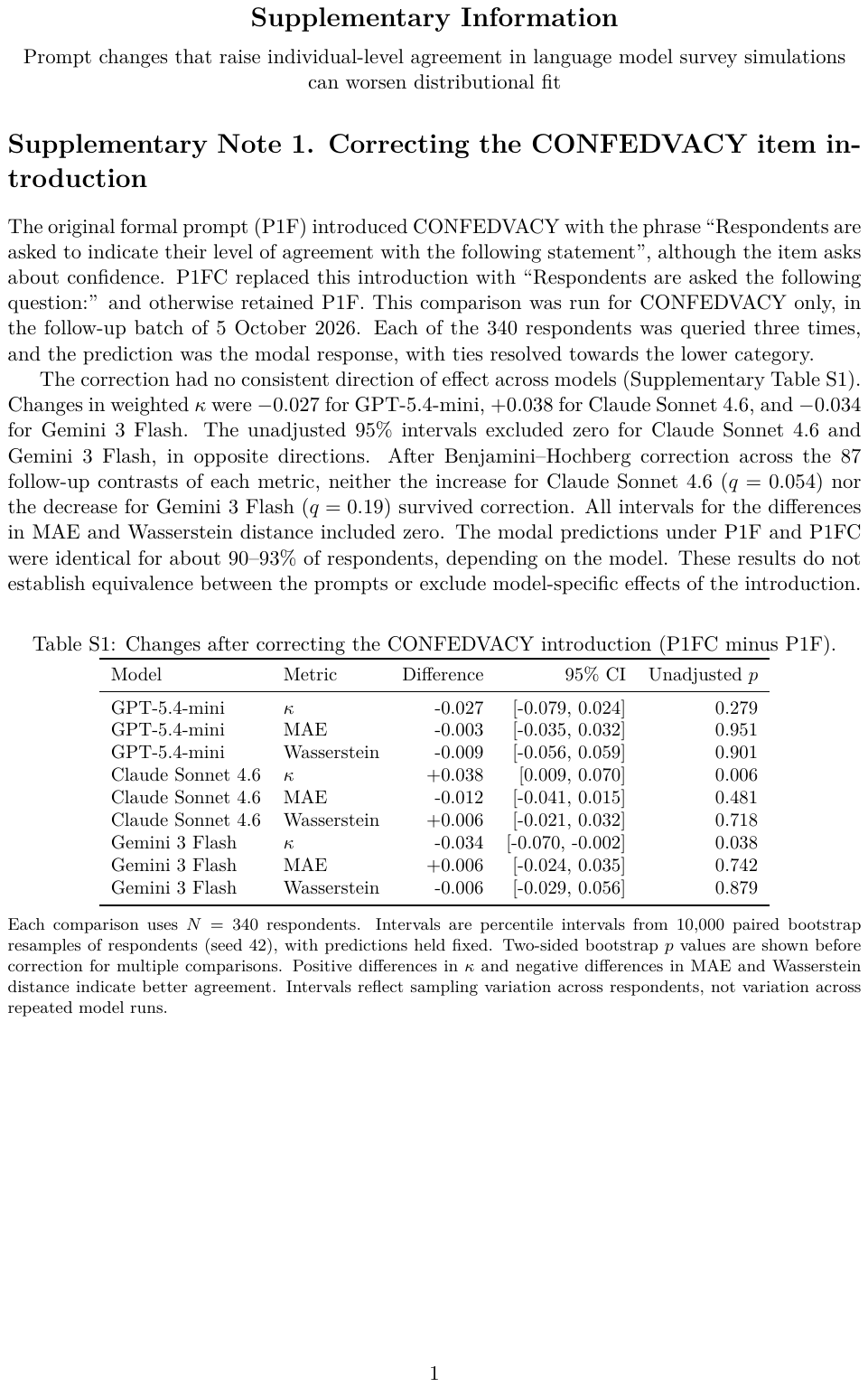}

\end{document}

%% file: sections/introduction.tex
Large language models (LLMs) are increasingly used in place of human survey respondents.
Given a person's demographic profile, a model can be asked how that person would answer a survey question, and the resulting ``silicon sample'' can then be compared with, or substituted for, answers from real respondents \cite{argyleOutOneMany2023,grossmannAITransformationSocial2023}.
Uses now extend well beyond opinion polling.
LLM simulations of nationally representative samples have predicted the results of survey experiments about as accurately as forecasts pooled across thousands of lay people \cite{ashokkumarLargeLanguageModels2026}.
A recent review counts silicon samples among four roles in which LLMs act as proxies for people, each requiring its own validity criteria \cite{karetnikovLargeLanguageModels2026}.
Health research has particular reason to take an interest, since surveys of health attitudes are expensive and slow to field.

How well these simulations represent individuals is a different matter.
When models predict individual answers from respondent profiles, simple statistical models using the same variables can do better, and the models' errors cluster in particular parts of the population \cite{pataranutapornSimulatingHumanWellbeing2025}.
Simulated samples can match the aggregate distribution of responses more closely than the answers of the individuals they stand for \cite{abeliukAuditingSociodemographicCrosssocietal2026}, and their accuracy varies across sociodemographic groups \cite{abeliukAuditingSociodemographicCrosssocietal2026,santurkarWhoseOpinionsLanguage2023}.
Results can also differ between models \cite{gaoTakeCautionUsing2025} and change over time for the same model \cite{bisbeeSyntheticReplacementsHuman2024}.
On science-related issues, models have misestimated public opinion in systematic directions \cite{leeCanLargeLanguage2024}.
Misestimation in a systematic direction is a particular concern for attitudes such as vaccine confidence, which in the United States divide sharply along political lines \cite{callaghanCorrelatesDisparitiesIntention2021,mottaPartisanConflictCOVID192023,baumgaertnerInfluencePoliticalIdeology2018}.

Results also depend on how the model is asked.
LLM outputs respond to features of a prompt that carry no substantive information, including formatting, the order in which information appears and the framing of response options \cite{sclarQuantifyingLanguageModels2024,brucksPromptArchitectureInduces2025,tjuatjaLLMsExhibitHumanlike2024}.
A survey simulation involves many such choices, and they are seldom justified.
The model can be addressed as the respondent or asked to report a respondent's answer, the profile can be written as prose or as a list, and politically charged attributes can appear early or late in it.
These choices matter.
In simulated COVID-19 vaccination decisions, third-person framing and the order of information shifted predicted vaccination rates, and the shifts were not the same across models \cite{xuComparingAlgorithmicFidelity2026}.
Failures to reproduce human behaviour have also varied unpredictably with the assigned role and the input language \cite{gaoTakeCautionUsing2025}.

What is still missing is evidence on which parts of a prompt effect hold across models when individual agreement and distributional fidelity are evaluated together.
Agreement with individual answers and fidelity to the distribution of answers are separate validation targets.
A prompt change can affect them differently, so an improvement on one need not carry over to the other.
Looking at both, across several models, makes it possible to identify effects that recur without assuming that a prompt validated on one model will behave the same way on another.

We examine this question with individual-level predictions of two vaccine attitudes in the 2022 General Social Survey (GSS), agreement that vaccines are safe (VAXSAFE) and confidence in federal oversight of COVID-19 vaccine safety (CONFEDVACY).
Three widely used models predicted the answers of 340 respondents from six sociodemographic variables under six prompts, which varied where party identification appeared in the profile and the style of the prompt.
We compared the predictions with the respondents' answers and with an ordered logit benchmark based on the same variables.
We consider prompt effects at two levels.
For individual agreement, measured by weighted $\kappa$ and mean absolute error, we ask whether a prompt change moves the predictions of different models in the same direction.
For the predicted distribution, measured by the Wasserstein distance, we ask whether a change improves or worsens fit consistently across models.
Under the base prompt, most models had lower weighted agreement than the benchmark, and prediction errors followed partisan and age lines.
Prompt changes raised weighted agreement in a consistent direction across models, but their effects on the predicted distribution were model-specific.
In a follow-up decomposition, the gain in agreement was associated mainly with presenting the profile as a structured list rather than with the formal tone of the prompt.

%% file: sections/results.tex
\subsection*{Consistency across repeated queries differed between models}

We first asked whether each model gave the same answer when the same respondent profile was submitted repeatedly.
Across 10 runs of the base prompt at temperature 0.2, Claude Sonnet 4.6 and Gemini 3 Flash were almost perfectly consistent, with Krippendorff's $\alpha$ of 0.970 and 0.972 for VAXSAFE and 0.976 and 0.984 for CONFEDVACY.
GPT-5.4-mini was less stable, particularly for VAXSAFE ($\alpha = 0.600$, against 0.769 for CONFEDVACY).
The primary analyses use the modal response across runs, so each respondent has a single prediction per model and prompt.
Ties were rare (Methods), but the mode itself could still differ between repeated sets of runs.
Intraclass correlations gave the same picture (Supplementary Table~S2).
The main distributional results did not depend on this choice.
All 24 prompt contrasts in the Wasserstein distance reported below retained the direction of the modal estimate both on average across 500 single-run draws and when responses were pooled across runs (Supplementary Note~10 and Supplementary Table~S19).

\subsection*{Base-prompt predictions mostly had lower weighted agreement than the benchmark}

\begin{figure}[tbp]
  \centering
  \includegraphics[width=\textwidth]{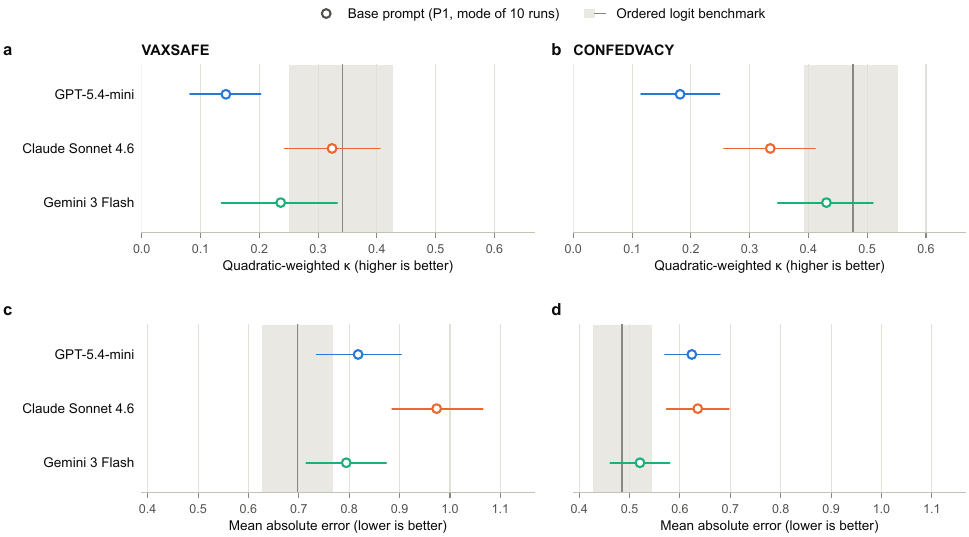}
  \caption{\textbf{Agreement between model predictions and observed responses under the base prompt, relative to a statistical benchmark.}
  \textbf{a,b}, Quadratic-weighted Cohen's $\kappa$ for VAXSAFE (\textbf{a}) and CONFEDVACY (\textbf{b}).
  \textbf{c,d}, Mean absolute error for VAXSAFE (\textbf{c}) and CONFEDVACY (\textbf{d}).
  VAXSAFE is agreement that vaccines are safe (5-point scale) and CONFEDVACY is confidence in federal oversight of vaccine safety (3-point scale).
  Circles show each model's predictions under the base prompt (P1), taken as the modal response across 10 runs.
  Grey vertical lines and bands show an ordered logit model fitted to the same six sociodemographic predictors and evaluated out of sample by repeated tenfold cross-validation.
  Points are estimates on the full analytic sample ($N = 340$).
  Error bars and bands are 95\% percentile bootstrap confidence intervals from 10,000 respondent-level resamples.
  Differences between a model and the benchmark were assessed with paired bootstrap intervals of the difference (Methods), not by the overlap of these intervals.}
  \label{fig:validity}
\end{figure}

We next compared each model's predictions under the base prompt with the answers the same 340 people actually gave, and with an ordered logit model that used the same six sociodemographic variables (Fig.~\ref{fig:validity}).
Agreement was modest throughout.
Quadratic-weighted $\kappa$ ranged from 0.14 (GPT-5.4-mini, VAXSAFE) to 0.43 (Gemini 3 Flash, CONFEDVACY).
The benchmark, evaluated out of sample by repeated tenfold cross-validation (Methods), reached 0.34 for VAXSAFE and 0.48 for CONFEDVACY.
In paired bootstrap comparisons, $\kappa$ was lower than the benchmark's in four of the six model--item combinations.
The differences ranged from $-0.11$ for Gemini 3 Flash on VAXSAFE (95\% confidence interval (CI) $-0.19$ to $-0.02$) to $-0.29$ for GPT-5.4-mini on CONFEDVACY (95\% CI $-0.37$ to $-0.22$).
For Claude Sonnet 4.6 on VAXSAFE ($-0.02$, 95\% CI $-0.08$ to $0.05$) and Gemini 3 Flash on CONFEDVACY ($-0.05$, $-0.11$ to $0.02$), $\kappa$ was statistically indistinguishable from the benchmark's.
Mean absolute error (MAE) separated the predictors more clearly.
It was larger than the benchmark's in five of the six combinations, and only for Gemini 3 Flash on CONFEDVACY was the difference indistinguishable from zero (0.52 against 0.49, difference $+0.04$, 95\% CI $-0.02$ to $+0.09$).

The two metrics did not always rank the models in the same order.
On VAXSAFE, Claude Sonnet 4.6 had the highest $\kappa$ of the three models (0.32) but also the largest error (MAE 0.97, against 0.70 for the benchmark).
Its $\kappa$ was statistically indistinguishable from the benchmark's even though its absolute errors were much larger (difference $+0.28$, 95\% CI $+0.19$ to $+0.37$).
The response distributions help explain why.

\subsection*{Models erred in opposite directions on vaccine safety}

\begin{figure}[tbp]
  \centering
  \includegraphics[width=\textwidth]{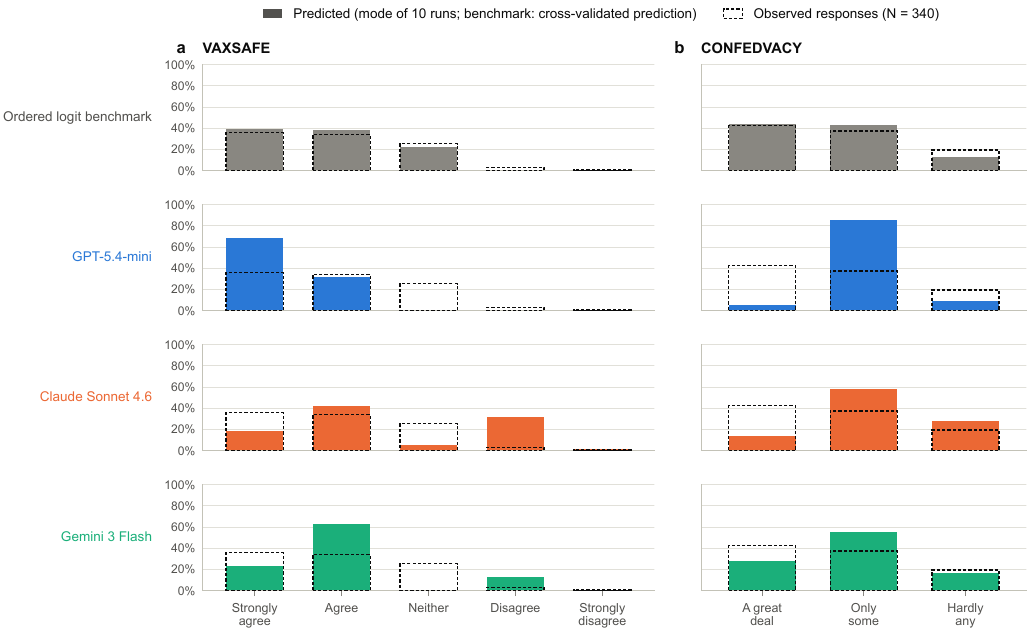}
  \caption{\textbf{Observed and predicted response distributions under the base prompt.}
  \textbf{a}, VAXSAFE, agreement that vaccines are safe (5-point scale).
  \textbf{b}, CONFEDVACY, confidence in federal oversight of vaccine safety (3-point scale).
  Each panel shows the share of the 340 respondents predicted to give each response, for the ordered logit benchmark (cross-validated predictions from the same six sociodemographic predictors) and for each model under the base prompt (P1, modal response across 10 runs).
  Dashed outlines show the observed distribution, repeated in every panel for reference.
  For the benchmark, GPT-5.4-mini, Claude Sonnet 4.6 and Gemini 3 Flash, in that order, the Wasserstein-1 distance between the predicted and observed distributions was 0.16, 0.67, 0.56 and 0.38 on VAXSAFE and 0.08, 0.48, 0.37 and 0.18 on CONFEDVACY.
  The variance ratio, the variance of the predictions divided by the variance of the observed responses, was 0.71, 0.26, 1.64 and 0.91 on VAXSAFE and 0.83, 0.24, 0.70 and 0.76 on CONFEDVACY.
  It is computed on the full analytic sample without confidence intervals.
  Values below 1 indicate predictions more concentrated than the observed responses, and values above 1 indicate predictions more dispersed.
  The benchmark, which uses the same six variables, shows how concentrated a conventional predictor's modal predictions are.}
  \label{fig:distributions}
\end{figure}

The response distributions show how the models departed from the observed data (Fig.~\ref{fig:distributions}).
For VAXSAFE, 70\% of respondents agreed or strongly agreed that vaccines are safe, 26\% neither agreed nor disagreed, and 4\% disagreed or strongly disagreed.
GPT-5.4-mini put every respondent in one of the two agreement categories and predicted that 68\% would agree strongly, which shifted its mean prediction 0.67 points towards agreement.
Claude Sonnet 4.6 went the other way.
It predicted that 34\% of respondents would disagree or strongly disagree, eight times the observed share, and its mean prediction lay 0.56 points towards disagreement.
Gemini 3 Flash was nearly unbiased on average (mean shift $+0.05$) but put 63\% of its predictions in a single category.
All three models almost never predicted the neutral category, although a quarter of respondents chose it.
GPT-5.4-mini assigned it to no one, Claude Sonnet 4.6 to 5\% of respondents and Gemini 3 Flash to 1\%.
The choice of model therefore decided not only how far the simulated population departed from the real one but also in which direction.

For CONFEDVACY, all three models moved respondents out of the most confident category and towards the middle option.
Although 43\% of respondents reported a great deal of confidence in federal oversight of vaccine safety, the models placed 5\% (GPT-5.4-mini), 14\% (Claude Sonnet 4.6) and 28\% (Gemini 3 Flash) of respondents there.
GPT-5.4-mini put 86\% of respondents in the middle category.
The Wasserstein distance between the predicted and observed distributions ranked the models in the same order on both items.
GPT-5.4-mini was furthest from the observed data (0.67 and 0.48), followed by Claude Sonnet 4.6 (0.56 and 0.37) and Gemini 3 Flash (0.38 and 0.18).
Only Gemini 3 Flash on CONFEDVACY came within sampling error of the benchmark on this measure (difference $+0.10$, 95\% CI $-0.003$ to $+0.165$).

We also looked at how widely each model spread its predictions, using the ratio of the variance of the predictions to that of the observed responses.
This comparison is descriptive.
It is not one of the design-based contrasts, and we computed no confidence intervals for it.
GPT-5.4-mini compressed its predictions to about a quarter of the observed variance on both items (variance ratios 0.26 and 0.24).
Claude Sonnet 4.6, by contrast, overdispersed its VAXSAFE predictions (1.64).
Gemini 3 Flash (0.91 and 0.76) came closest to the benchmark (0.71 and 0.83).
The benchmark, which uses the same six variables, also compressed the observed variance, so compression by itself does not set the models apart from a conventional predictor.
GPT-5.4-mini's compression went well beyond the benchmark's, in line with reports that LLM answers vary less than human answers do, including across sociodemographic profiles \cite{boelaertMachineBiasHow2025,bisbeeSyntheticReplacementsHuman2024}.
Taking the mode did not produce this compression.
With one randomly drawn run per respondent, GPT-5.4-mini's variance ratios were 0.27 on both items.
Together with its lower run-to-run consistency, this compression calls for caution in interpreting the agreement metrics of GPT-5.4-mini.
Full distributional statistics, including boundary avoidance, are reported in Supplementary Table~S8.

\subsection*{Prediction errors varied with party identification and age}

\begin{figure}[tbp]
  \centering
  \includegraphics[width=\textwidth]{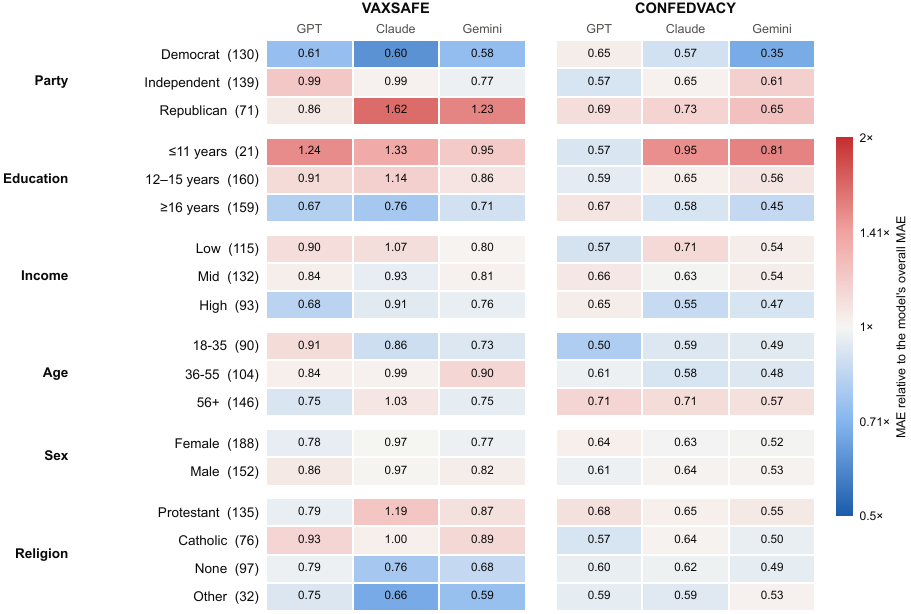}
  \caption{\textbf{Prediction error across sociodemographic subgroups under the base prompt.}
  Each cell gives the mean absolute error (MAE) of one model for one item within one subgroup.
  Subgroup sizes are shown in parentheses.
  Cell colour shows the subgroup MAE relative to that model's overall MAE on the same item, on a log scale from half (blue) to twice (red) the overall value.
  Predictions are modal responses across 10 runs.
  Column labels GPT, Claude and Gemini refer to GPT-5.4-mini, Claude Sonnet 4.6 and Gemini 3 Flash.}
  \label{fig:subgroups}
\end{figure}

Errors were not spread evenly across respondents (Fig.~\ref{fig:subgroups}).
For VAXSAFE, Claude Sonnet 4.6 and Gemini 3 Flash were least accurate for Republicans, whereas GPT-5.4-mini was least accurate for Independents (MAE 0.99).
Every model was less accurate for Republicans than for Democrats.
MAE for Republicans was 0.86 (GPT-5.4-mini), 1.62 (Claude Sonnet 4.6) and 1.23 (Gemini 3 Flash), against 0.61, 0.60 and 0.58 for Democrats, a partisan gap ranging from a factor of 1.4 for GPT-5.4-mini to 2.7 for Claude Sonnet 4.6.
VAXSAFE errors also fell with education in all three models, although the least educated group was small ($n = 21$).
On CONFEDVACY the differences between groups were smaller.
The most consistent pattern there concerned age, with respondents aged 56 and over having the largest errors in all three models (MAE 0.71, 0.71 and 0.57).
Mean errors were similar for women and men.

These characteristics are correlated, so we also regressed each model's absolute error on all six variables at once (Table~\ref{tab:errors}).
For VAXSAFE, party identification was the only characteristic associated with larger errors in every model.
Each step towards the Republican end of the seven-point scale added 0.05 (GPT-5.4-mini), 0.19 (Claude Sonnet 4.6) and 0.12 (Gemini 3 Flash) to the expected absolute error.
Over the full scale, these increments come to about 0.3, 1.2 and 0.7 response categories.
More years of education went with smaller errors for GPT-5.4-mini and Claude Sonnet 4.6.
The six variables together accounted for 28\% of the variation in Claude Sonnet 4.6's VAXSAFE errors, compared with 13\% for Gemini 3 Flash and 7\% for GPT-5.4-mini.
The partisan gradient was thus steepest for the model whose predictions were shifted furthest towards scepticism.
For CONFEDVACY the variables explained little (4 to 7\%).
Errors rose with age for GPT-5.4-mini (0.05 per decade) and Claude Sonnet 4.6 (0.04 per decade), rose with Republican identification for Gemini 3 Flash (0.06 per step), and fell with education for Claude Sonnet 4.6.
Once the other characteristics were held constant, no coefficient for sex, family income or religious group was distinguishable from zero in any model.

\begin{table}[tbp]
  \centering
  \small
  \caption{\textbf{Respondent characteristics associated with absolute
  prediction error under the base prompt.} Ordinary least squares regressions of
  the absolute difference between the predicted and observed response on the six
  sociodemographic variables, estimated separately for each model and item
  ($N = 340$). Entries are coefficients with heteroskedasticity-consistent (HC3)
  standard errors in parentheses.
  Positive values indicate larger errors. Party identification runs from strong
  Democrat (0) to strong Republican (6). Family income is CONINC, in constant 2000
  dollars. The reference category for religion is
  no religion. $^{*}P < 0.05$, $^{**}P < 0.01$, $^{***}P < 0.001$ (two-sided).}
  \label{tab:errors}
  \begin{tabular}{lccc}
    \toprule
     & GPT-5.4-mini & Claude Sonnet 4.6 & Gemini 3 Flash \\
    \midrule
    \multicolumn{4}{l}{\textit{VAXSAFE}} \\
    Age (per 10 years) & $-0.043$ (0.023) & $0.031$ (0.023) & $-0.022$ (0.022) \\
    Female & $-0.055$ (0.083) & $0.113$ (0.083) & $0.016$ (0.080) \\
    Education (years) & $-0.045^{**}$ (0.017) & $-0.041^{*}$ (0.018) & $-0.013$ (0.016) \\
    Family income (per \$10,000, 2000 dollars) & $-0.003$ (0.011) & $-0.011$ (0.009) & $-0.001$ (0.010) \\
    Party identification (0--6) & $0.051^{*}$ (0.023) & $0.192^{***}$ (0.024) & $0.122^{***}$ (0.025) \\
    Protestant & $-0.047$ (0.104) & $0.143$ (0.112) & $0.057$ (0.094) \\
    Catholic & $0.096$ (0.111) & $0.051$ (0.119) & $0.127$ (0.105) \\
    Other religion & $-0.014$ (0.169) & $-0.096$ (0.128) & $-0.073$ (0.152) \\
    $R^2$ & 0.073 & 0.282 & 0.131 \\
    \midrule
    \multicolumn{4}{l}{\textit{CONFEDVACY}} \\
    Age (per 10 years) & $0.047^{**}$ (0.016) & $0.039^{*}$ (0.018) & $0.017$ (0.017) \\
    Female & $0.047$ (0.056) & $0.006$ (0.064) & $0.030$ (0.061) \\
    Education (years) & $0.006$ (0.011) & $-0.028^{*}$ (0.012) & $-0.017$ (0.012) \\
    Family income (per \$10,000, 2000 dollars) & $0.001$ (0.007) & $-0.009$ (0.007) & $-0.007$ (0.007) \\
    Party identification (0--6) & $0.008$ (0.015) & $0.027$ (0.017) & $0.063^{***}$ (0.016) \\
    Protestant & $0.028$ (0.071) & $-0.051$ (0.083) & $-0.046$ (0.080) \\
    Catholic & $-0.065$ (0.079) & $-0.035$ (0.088) & $-0.061$ (0.086) \\
    Other religion & $-0.027$ (0.104) & $-0.031$ (0.112) & $0.039$ (0.110) \\
    $R^2$ & 0.038 & 0.051 & 0.068 \\
    \bottomrule
  \end{tabular}
\end{table}

\subsection*{Weighted agreement was higher in 23 of 24 contrasts, seven after correction}

We then asked how two features of the prompt changed the predictions.
One was where party identification appeared in the profile (last, first or in the middle).
The other was whether the prompt was written in a casual, second-person style or as a formal survey record (Methods).
Four design-based contrasts summarize this $3 \times 2$ design, each estimated within model and item on the same respondents (Fig.~\ref{fig:tradeoff} and Supplementary Fig.~S2).
Across these contrasts, 23 of the 24 differences in $\kappa$ were positive, and none of the 48 contrasts in $\kappa$ or MAE showed a deterioration distinguishable from zero.
Ten $\kappa$ differences had intervals that excluded zero, eight of them on CONFEDVACY.
Seven of these ten survived false discovery rate correction, and five after Bonferroni correction (Supplementary Table~S9).

The clearest result came from combining both changes.
Placing party identification first in a formal survey-style prompt (P2F versus P1) increased $\kappa$ for CONFEDVACY in all three models, by $+0.180$ for GPT-5.4-mini (95\% CI 0.102 to 0.262), $+0.134$ for Claude Sonnet 4.6 (0.078 to 0.192) and $+0.073$ for Gemini 3 Flash (0.013 to 0.134).
The gains for GPT-5.4-mini and Claude Sonnet 4.6 survived false discovery rate correction, whereas the gain for Gemini 3 Flash did not ($q = 0.054$).
For GPT-5.4-mini and Claude Sonnet 4.6, most of this gain came from the position of party identification alone (P2 versus P1, $+0.161$ and $+0.122$), which also reduced their MAE ($-0.079$ and $-0.094$).
For Gemini 3 Flash, neither the position effect ($+0.017$, 95\% CI $-0.028$ to $0.062$) nor the formal prompt on its own was distinguishable from zero, although the two together raised $\kappa$.
Their interaction was not distinguishable from zero ($+0.018$, 95\% CI $-0.045$ to $0.081$, Supplementary Table~S13).
The formal survey-style prompt by itself raised CONFEDVACY $\kappa$ for Claude Sonnet 4.6 (P1F versus P1, $+0.078$, 0.032 to 0.125) and, averaged over the three party positions, for GPT-5.4-mini ($+0.081$, 0.027 to 0.136).
On VAXSAFE the changes in $\kappa$ under the formal prompt were positive in all three models but not distinguishable from zero (for example, GPT-5.4-mini $+0.080$, 95\% CI $-0.022$ to $0.175$).
The formal prompt's clearest effect on VAXSAFE was a lower MAE for Claude Sonnet 4.6 ($-0.126$, $-0.191$ to $-0.062$).

A higher weighted $\kappa$ does not always mean smaller errors.
The statistic compares the mean squared error with the error expected from the spread and the means of the two sets of responses, so it can rise when predictions become more dispersed even if their errors do not shrink.
We therefore split each change in $\kappa$ into a part due to the change in mean squared error and a part due to the change in the spread and mean of the predictions (Methods and Supplementary Table~S18).
This decomposition is descriptive and uses point estimates.
Of the 23 contrasts with a positive change in $\kappa$, 8 had a higher point estimate of mean squared error, so their gain in $\kappa$ came from more dispersed predictions rather than from smaller errors.
Three of these were for Gemini 3 Flash on CONFEDVACY.
Under the formal prompt alone, the formal average and the combined change, its mean squared error was higher by 0.029, 0.055 and 0.047, and its MAE under the combined change was unchanged ($0.000$, 95\% CI $-0.056$ to $0.056$).
For Gemini 3 Flash on this item, the gains in $\kappa$ therefore came without evidence of smaller errors.
The contrast averaged over party positions for Claude Sonnet 4.6 on CONFEDVACY, which supports our main distributional finding, was also among the eight.
Its mean squared error was higher by 0.017, so its gain of 0.046 in $\kappa$ came from the spread and mean of its predictions rather than from smaller errors.
Under the formal prompt alone, by contrast, Claude Sonnet 4.6 lowered its MAE on CONFEDVACY ($-0.044$, 95\% CI $-0.082$ to $-0.009$), although this change narrowly did not survive false discovery rate correction ($q = 0.054$), and its lower mean squared error accounted for 0.033 of its gain of 0.078 in $\kappa$.

\subsection*{Effects on the predicted distribution depended on the model}

\begin{figure}[tbp]
  \centering
  \includegraphics[width=\textwidth]{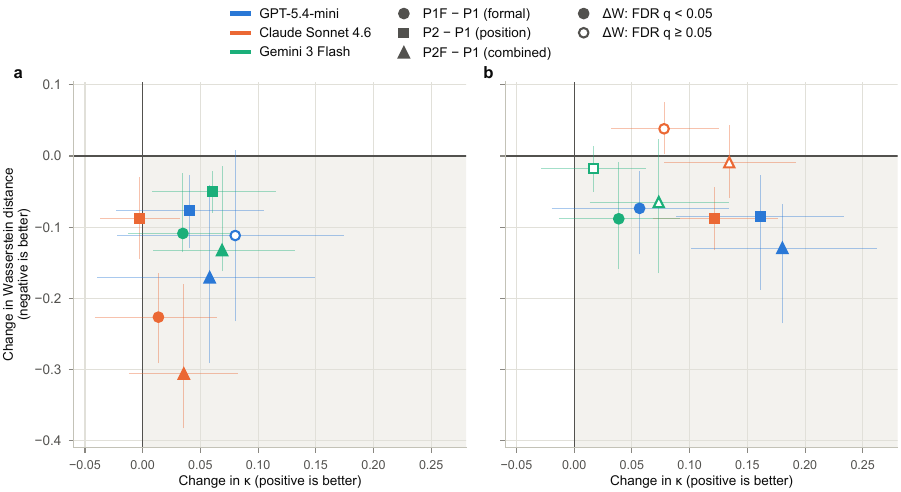}
  \caption{\textbf{Changes in individual-level agreement and in distributional fit under the same prompt changes.}
  \textbf{a}, VAXSAFE.
  \textbf{b}, CONFEDVACY.
  Each point is one model under one prompt change, estimated within model and item on the same respondents.
  The horizontal axis is the change in quadratic-weighted $\kappa$, where positive values indicate higher weighted agreement.
  The vertical axis is the change in the Wasserstein-1 distance between the predicted and observed response distributions, where negative values indicate a predicted distribution closer to the observed one.
  Points in the shaded lower right quadrant therefore improved on both levels, and points in the upper right quadrant raised weighted agreement while moving the predicted distribution further from the observed one.
  Colours identify models and marker shapes identify the prompt change: the formal survey-style prompt with party identification last (P1F minus P1), moving party identification first (P2 minus P1) and both changes together (P2F minus P1).
  The contrast averaged over party positions is not shown because it overlaps with these contrasts.
  Its values are given in Supplementary Table~S9.
  Thin horizontal and vertical lines are 95\% paired percentile bootstrap confidence intervals for the two changes, from 10,000 respondent-level resamples and not adjusted for multiple comparisons.
  Filled markers indicate a change in the Wasserstein distance with a Benjamini--Hochberg $q$ below 0.05 within the 24 contrasts for that metric.
  $N = 340$.}
  \label{fig:tradeoff}
\end{figure}

The same prompt changes also moved the predicted distributions, and here the models parted ways (Fig.~\ref{fig:tradeoff} and Supplementary Fig.~S2e,f).
Of the 24 contrasts in the Wasserstein distance, 17 had intervals that excluded zero.
Of these 17, 15 were decreases, and 16 survived false discovery rate correction.
For VAXSAFE, where no change in $\kappa$ survived correction, the distributional effects were clearer.
Combining both changes brought the predicted distribution closer to the observed one in every model (GPT-5.4-mini $-0.171$, Claude Sonnet 4.6 $-0.306$, Gemini 3 Flash $-0.132$), and the formal prompt alone did so for Claude Sonnet 4.6 ($-0.226$) and Gemini 3 Flash ($-0.109$).
On this item, prompt changes gave clearer evidence of better distributional fit than of better individual-level agreement.

For CONFEDVACY, even the direction of the distributional change differed between models.
The formal prompt moved the predicted distribution closer to the observed one for GPT-5.4-mini ($-0.074$) and Gemini 3 Flash ($-0.088$) but further from it for Claude Sonnet 4.6 ($+0.038$, 95\% CI 0.003 to 0.076, and $+0.036$, 0.008 to 0.066, when averaged over party positions), even though the same change raised Claude Sonnet 4.6's $\kappa$.
These were the only two contrasts whose intervals indicated a worse distribution.
The averaged contrast survived false discovery rate correction ($q = 0.023$), whereas the contrast with party identification last narrowly did not ($q = 0.058$).
This separation did not depend on taking the mode.
With one randomly drawn run per respondent, the formal prompt raised Claude Sonnet 4.6's $\kappa$ by 0.062 on average and increased its distance by 0.047, and pooling all runs gave the same increase in distance (95\% CI 0.013 to 0.082).
Moving party identification first narrowed the distance for GPT-5.4-mini and Claude Sonnet 4.6 ($-0.085$ and $-0.088$, both surviving false discovery rate correction), whereas the estimate for Gemini 3 Flash was smaller ($-0.018$) and did not survive correction.
Under the combined change, the estimated distance decreased in all three models (GPT-5.4-mini $-0.129$, Claude Sonnet 4.6 $-0.009$, Gemini 3 Flash $-0.065$), but only the decrease for GPT-5.4-mini survived correction.
Changes that raised weighted agreement in a common direction therefore left each model's predicted distribution in a different place, which is consistent with the models starting from differently biased distributions (Fig.~\ref{fig:distributions}).

A difference that is significant in one model but not in another does not by itself show that the two models differ, so we also compared the models directly.
For each contrast, we took the difference between two models' changes and estimated it on the same bootstrap resamples (Methods and Supplementary Table~S20).
For the formal prompt on CONFEDVACY, Claude Sonnet 4.6's change in the Wasserstein distance differed from that of GPT-5.4-mini ($0.112$, 95\% CI 0.041 to 0.194) and from that of Gemini 3 Flash ($0.126$, 0.032 to 0.215), and both differences survived false discovery rate correction.
The corresponding differences in $\kappa$ were not distinguishable from zero ($0.021$ and $0.039$), although this does not show that the gains in $\kappa$ were equal.
Across the 24 between-model comparisons for each metric, 9 differences in the Wasserstein distance survived correction, against 2 in $\kappa$.
Both of the latter reflected the smaller effect of party position for Gemini 3 Flash on CONFEDVACY.

\subsection*{Gains in agreement were mainly associated with the structured profile}

\begin{figure}[tbp]
  \centering
  \includegraphics[width=\textwidth]{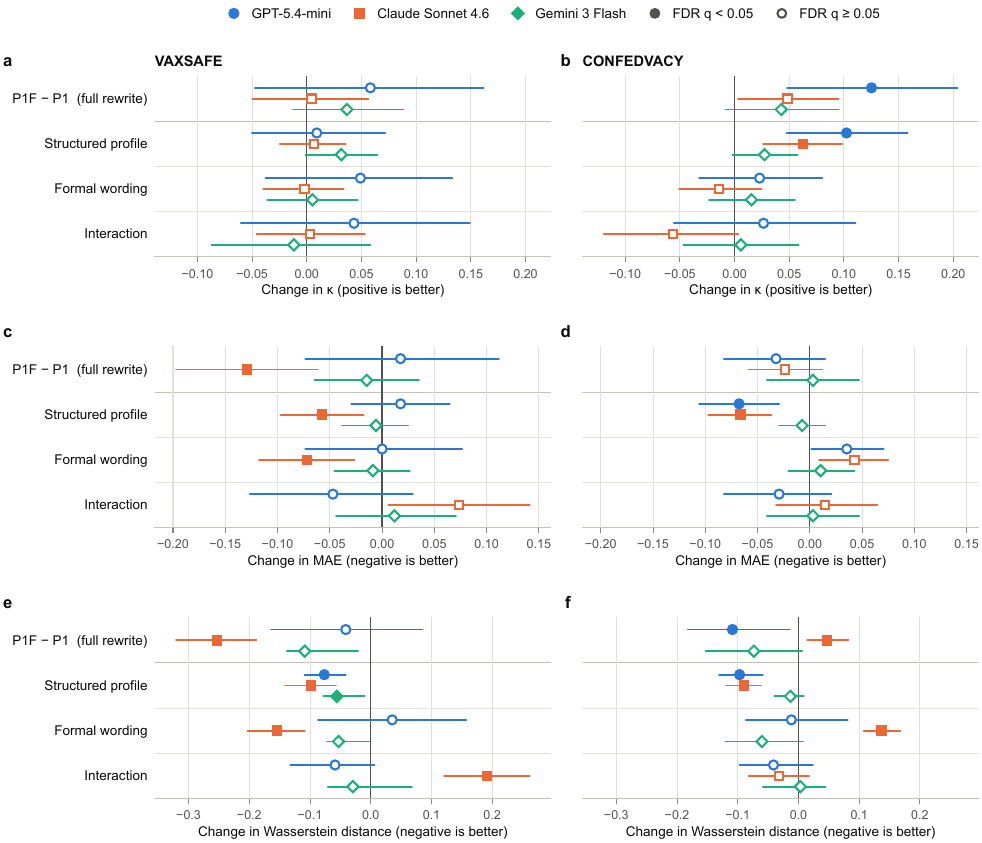}
  \caption{\textbf{Follow-up decomposition of the formal survey-style prompt.}
  In a separate batch collected after the main experiment, the two components of the formal rewrite were crossed in a $2 \times 2$ design.
  The components were profile layout (sentences, as in P1, versus a structured list of fields, as in P1F) and wording (the casual, second-person wording of P1 versus the formal, third-person wording of P1F), with party identification last throughout.
  The top row in each panel is the full rewrite (P1F minus P1) re-estimated in this batch, so its values differ from those of the main experiment (Fig.~\ref{fig:tradeoff} and Supplementary Fig.~S2).
  Structured profile is the effect of the list layout averaged over the two wordings, and formal wording is the effect of the formal wording averaged over the two layouts.
  The two effects sum to the full rewrite.
  Interaction is the difference between the layout effect under formal and under casual wording.
  \textbf{a,b}, Change in quadratic-weighted $\kappa$ for VAXSAFE (\textbf{a}) and CONFEDVACY (\textbf{b}).
  Positive values indicate higher weighted agreement.
  \textbf{c,d}, Change in mean absolute error.
  Negative values indicate smaller error.
  \textbf{e,f}, Change in the Wasserstein-1 distance between the predicted and observed response distributions.
  Negative values indicate a predicted distribution closer to the observed one.
  Predictions are modal responses across 3 runs.
  Points are differences on the full analytic sample ($N = 340$).
  Error bars are 95\% paired percentile bootstrap confidence intervals from 10,000 respondent-level resamples, not adjusted for multiple comparisons.
  Filled markers indicate differences with a Benjamini--Hochberg $q$ below 0.05 within each metric's family of 87 follow-up contrasts (Supplementary Table~S12).}
  \label{fig:decomposition}
\end{figure}

The formal survey-style prompt changed more than tone.
It also addressed the model in the third rather than the second person, and it listed the profile as labelled fields instead of sentences (Methods).
In a follow-up decomposition run after the main experiment, we crossed profile layout (sentences or a structured list) with wording (the casual, second-person wording of P1 or the formal, third-person wording of P1F), keeping party identification last (Fig.~\ref{fig:decomposition}).
In this batch the full rewrite raised CONFEDVACY $\kappa$ for GPT-5.4-mini ($+0.125$) and, less clearly, for Claude Sonnet 4.6 ($+0.048$, which did not survive correction).

The decomposition associates these gains mainly with the structured profile, which raised CONFEDVACY $\kappa$ by 0.102 for GPT-5.4-mini and 0.063 for Claude Sonnet 4.6.
Both effects survived false discovery rate correction.
The corresponding wording effects were small and inconsistent in direction ($+0.023$ and $-0.014$), and no wording effect on $\kappa$ was distinguishable from zero for any model or item.
MAE pointed the same way.
On CONFEDVACY the structured profile lowered the error for both models ($-0.068$ and $-0.066$), while the formal wording raised it slightly ($+0.035$ and $+0.043$, neither surviving correction).
The Wasserstein distance was lower with the structured profile in all six model--item combinations, and the decrease survived correction in five.
The wording effect on the distribution varied by model.
For Claude Sonnet 4.6 on CONFEDVACY, the formal wording, which also carried the third-person perspective and the agreement framing of the original item introduction, moved the distribution away from the observed one ($+0.137$) while the structured layout moved it closer ($-0.090$), which reproduces the adverse net distributional effect in this batch.
For Claude Sonnet 4.6 on VAXSAFE, the two components were not additive when each was introduced separately from P1.
Either change on its own reduced the distance by about as much as both together (structured layout alone $-0.194$, formal wording alone $-0.250$, both $-0.253$).
The interaction ($+0.191$) survived correction.

These results suggest one reading of the CONFEDVACY gains, which we offer as a post hoc observation.
In GPT-5.4-mini and Claude Sonnet 4.6, placing party identification first and listing it as a separate field may do similar work, since both make it more prominent in the profile.
In line with this, the formal prompt added less for Claude Sonnet 4.6 in the main experiment when party identification already came first (difference between the two formal effects $-0.065$, 95\% CI $-0.129$ to $-0.002$, uncorrected).

The formal effects were not tied to one wording.
We wrote two further formal versions with different phrasing.
Before running them, we set the criterion that the changes in $\kappa$ and in the Wasserstein distance should have the same sign as those of the original formal prompt in at least four of the six model--item combinations.
Both versions met it (Supplementary Fig.~S1).
Across the three formal wordings, 17 of 18 changes in $\kappa$ were positive and 14 of 18 Wasserstein distances fell.
The Claude Sonnet 4.6 exception on CONFEDVACY appeared under all three wordings.
After false discovery rate correction, the two alternative wordings differed from the original wording in 1 of 12 comparisons for $\kappa$, 6 of 12 for the Wasserstein distance and 3 of 12 for MAE.
These counts do not establish that the wordings are equivalent.

We also tested grammatical perspective more directly, by addressing the model in the second rather than the third person within one formal wording (P1F4 versus P1F3, which also differed in the tense of the opening clause).
Its effect on $\kappa$ was not distinguishable from zero in any model--item combination (all differences smaller than 0.04 in absolute value).
Where perspective did matter, the third-person version did worse.
It increased MAE for Gemini 3 Flash on VAXSAFE ($+0.129$) and Claude Sonnet 4.6 on CONFEDVACY ($+0.059$), and it moved the predicted distribution further from the observed one for GPT-5.4-mini on VAXSAFE ($+0.065$) and for Claude Sonnet 4.6 ($+0.112$) and Gemini 3 Flash ($+0.041$) on CONFEDVACY.
All five differences survived false discovery rate correction.
As in simulations of vaccination decisions \cite{xuComparingAlgorithmicFidelity2026}, the framing of whose answer the model gives shifted its predictions, but here it did so without a detectable change in $\kappa$.

Correcting a misworded introduction to the CONFEDVACY item in the original formal prompt changed most metrics little.
The exception was $\kappa$, which rose for Claude Sonnet 4.6 ($+0.038$) and fell for Gemini 3 Flash ($-0.034$).
Both intervals excluded zero, but neither change survived false discovery rate correction ($q = 0.054$ and $0.19$).
Details are in Supplementary Note~1 and Supplementary Table~S1.

Re-running the base and original formal prompts one to two months after the main experiment changed no model's $\kappa$ by more than 0.043, but it did move some predicted distributions.
Under the re-run base prompt, the Wasserstein distance on VAXSAFE fell by 0.085 for GPT-5.4-mini (95\% CI $-0.132$ to $-0.038$) and rose by 0.044 for Claude Sonnet 4.6 (0.006 to 0.079), and both changes survived false discovery rate correction (Supplementary Table~S12).
For Claude Sonnet 4.6 on CONFEDVACY, the estimated distance from the observed distribution changed by 0.000 under the base prompt and by $+0.009$ under the formal prompt.
Between batches, the GPT-5.4-mini formal effect on CONFEDVACY grew from $+0.057$ to $+0.125$ in $\kappa$, so effects seen only in that model should be read with caution.

A separate check pointed the same way.
Family income had been presented as CONINC, in constant 2000 dollars, without stating the base year.
In a further batch, we replaced it with the questionnaire's income bracket or stated the base year, keeping everything else fixed (Methods and Supplementary Note~9).
No change in $\kappa$ survived false discovery rate correction, all $\kappa$ differences were smaller than 0.07 in absolute value, and 80--95\% of modal predictions stayed identical.
The predicted distributions moved more.
On VAXSAFE, presenting income as a bracket in the casual prompt moved GPT-5.4-mini's distribution further from the observed one ($+0.118$, 95\% CI 0.071 to 0.162) and Claude Sonnet 4.6's closer to it ($-0.132$, $-0.185$ to $-0.079$), and both changes survived correction.
This check was not designed to test our main comparison, but it shows the same pattern in a separate manipulation.
A minor change to the profile moved the estimated VAXSAFE distance in opposite directions for GPT-5.4-mini and Claude Sonnet 4.6, without a detectable change in $\kappa$.

%% file: sections/discussion.tex
Prompt changes can raise weighted agreement with individual responses without improving the distribution of simulated responses.
In all three models studied here, placing party identification first in a formal survey-style prompt increased weighted agreement with respondents' confidence in federal oversight, and the changes for GPT-5.4-mini and Claude Sonnet 4.6 survived false discovery rate correction.
For Gemini 3 Flash, the gain did not survive correction and reflected more dispersed predictions, with no evidence of smaller errors.
More broadly, 23 of 24 design-based contrasts in $\kappa$ were positive, although many were not distinguishable from zero.
Distributional effects were less consistent across models.
Averaged across party positions, for instance, formal prompts raised Claude Sonnet 4.6's weighted agreement on CONFEDVACY while moving its predicted distribution further from the observed one, and this distributional shift differed directly from those of the other two models.
With party identification last, the same shift appeared under all three formal wordings.
These results distinguish two levels of prompt evaluation.
A prompt change can move individual-level agreement in the same direction across models and still have model-specific consequences for distributional fidelity.

The distinction concerns the changes a prompt induces, not simply whether individual predictions or aggregate distributions are more accurate.
Previous research has evaluated both individual and aggregate fidelity \cite{abeliukAuditingSociodemographicCrosssocietal2026,pataranutapornSimulatingHumanWellbeing2025}, and simulations of vaccination decisions have shown model-specific changes in predicted uptake under different prompt framings, including third-person framing, and information orders \cite{xuComparingAlgorithmicFidelity2026}.
We bring these strands together by examining the same prompt contrasts at both levels.
The results show why a prompt cannot be judged successful from higher individual-level agreement alone.
A prompt that raises agreement on one model may be a useful starting point on another, but its distributional consequences still need to be evaluated separately.

Why the clearest agreement gains appeared for confidence in federal oversight rather than for vaccine safety is an open question.
One possibility is that prompt changes help when they make an informative attribute more prominent.
In our unweighted sample, 70.0\% of Democrats and 18.3\% of Republicans reported a great deal of confidence in federal oversight of vaccine safety, while 53.1\% of Democrats and 22.5\% of Republicans strongly agreed that vaccines are safe.
The mean partisan differences were 0.85 on the three-point CONFEDVACY scale and 0.64 on the five-point VAXSAFE scale, roughly 43\% and 16\% of the respective scale ranges.
These figures are descriptive comparisons between different ordinal items, but they are consistent with greater partisan separation on CONFEDVACY in this sample.
Moving party identification to the front of the profile and displaying it as a separate field could both make it more prominent.
Their partly overlapping effects in GPT-5.4-mini and Claude Sonnet 4.6 are consistent with this interpretation, which remains post hoc.
Evidence that the position of information affects LLM performance \cite{liuLostMiddleHow2024} offers a possible analogy, not a direct test.
Neither the partisan differences nor our prompt comparisons show which information the models used internally.

Model-specific error patterns also argue for separate validation.
The six sociodemographic variables explained 28\% of the variation in Claude Sonnet 4.6's VAXSAFE errors but 7\% of GPT-5.4-mini's, and the two models erred in opposite directions.
GPT-5.4-mini predicted universal agreement that vaccines are safe, whereas Claude Sonnet 4.6 predicted substantially more scepticism than respondents expressed.
Work on global-warming opinions has suggested that prioritizing factual correctness may make it harder to simulate people who do not share scientifically supported beliefs \cite{leeCanLargeLanguage2024}.
That possibility does not explain both directions of bias here, and our design cannot attribute the differences to particular training or alignment choices.
For VAXSAFE, all three models were less accurate for Republican than for Democratic respondents, and Republicans were the least accurately predicted group for two of them.
Systematic errors can thus concentrate within a politically relevant subgroup.
Which groups are hardest to simulate also differs across studies.
In the US sample of one cross-national study, older respondents were simulated better \cite{quPerformanceBiasesLarge2024}.
In another, accuracy fell with age, most clearly in Chile \cite{abeliukAuditingSociodemographicCrosssocietal2026}.
Americans aged 65 and over were among the groups whose opinions language models reflected least well \cite{santurkarWhoseOpinionsLanguage2023}, and Generation Z was among the identity groups that models most often misportrayed \cite{wangLargeLanguageModels2025a}.
Subgroup audits therefore need to be specific to the model, question and population under study.

Our prediction target was the answer each respondent actually selected, not the full set of considerations behind it.
People may hold conflicting considerations or be unsure between adjacent categories, yet a survey records one response at the time of asking \cite{zallerSimpleTheorySurvey1992}.
Predicting that recorded response is therefore a meaningful task even when the underlying attitude is neither fixed nor internally uniform.
We evaluated one categorical prediction per respondent, the mode of repeated model outputs.
This aggregation is a separate choice from the prediction target, and a modal model output need not reproduce how a person resolves uncertainty on a particular occasion.
In our data, however, this choice did not drive the main patterns.
All 24 main Wasserstein contrasts kept their direction both on average across 500 single-run draws, each corresponding to querying every profile once, and when responses were pooled across runs.
In the single-run analysis, GPT-5.4-mini's base-prompt variance ratio was 0.27 on both items, close to the values under the mode, so its compression was not produced by taking the mode.
Our conclusions therefore do not depend on the mode, but they do not extend to other ways of eliciting responses, such as asking a model to report a distribution directly.

The near absence of neutral predictions on VAXSAFE does not reflect a general reluctance to choose the middle of a scale.
On CONFEDVACY, the models placed between 56\% and 86\% of respondents in the middle category, whereas 38\% of respondents chose it.
What the models avoided on VAXSAFE was a neutral option, which survey respondents often use to signal that they have no opinion rather than a middle position.
In an earlier survey study that probed such answers, most respondents who chose ``neither agree nor disagree'' turned out to have no opinion on the issue, and the option served partly as a face-saving way of saying ``don't know'' \cite{sturgisMiddleAlternativesRevisited2014}.
If some of the 26\% of respondents who chose the neutral category on VAXSAFE used it in this way, the models' predictions are consistent with inferring a position from each profile rather than representing the absence of one.
A recent comparison of human and synthetic survey respondents points in the same direction.
Human respondents selected ``not sure'' about four times as often as synthetic respondents did on opinion questions that offered that option (16\% versus 4\%) \cite{chapekisCanAIStand}.
Two caveats apply.
The evidence that neutral responses often mask the absence of an opinion comes mainly from issues of low public salience, whereas vaccine safety was widely discussed in 2022, so such responses may make up a smaller share of our neutral respondents.
Our data also cannot show what any individual meant by a neutral answer, and this interpretation is post hoc.
If so, silicon samples may misrepresent not only which positions people hold but also how many people hold no position at all, a possibility that data on respondents' reasons would be needed to test.

Validation also has a temporal dimension.
When the base and formal prompts were re-run one to two months after the main experiment, the formal effect on CONFEDVACY for GPT-5.4-mini rose from 0.057 to 0.125 in $\kappa$, and the re-run base prompt moved the predicted VAXSAFE distribution in opposite directions for GPT-5.4-mini and Claude Sonnet 4.6.
Prompt effects may therefore be version-dependent as well as model-dependent, and validation should not be assumed to carry over across access dates or model versions, even under the same model identifier \cite{bisbeeSyntheticReplacementsHuman2024}.
Version changes are one possible explanation, but the comparison cannot separate provider updates from variation across runs.
Request configuration may also matter.
In an exploratory pilot with 50 respondents, Gemini 3 Flash showed higher agreement under the API default configuration than with \texttt{thinking\_budget=0} (Supplementary Note~4 and Table~S4).
The two configurations were evaluated almost five months apart, so this descriptive comparison cannot isolate an effect of reasoning from temporal changes or variation across runs.
The main results apply to the reported configurations, not to every way these models can be deployed.

Several limitations apply.
The analytic sample comprised 340 respondents with complete data.
Participation in the web-only GSS follow-on was selective, with a 38\% retention rate, and the AmeriSpeak oversample was not eligible.
All distributional comparisons describe this unweighted sample, not population-representative estimates.
We studied two items in one country with three models and six prompts.
The best prompts were selected on the same data, so their comparison with the benchmark is optimistic for the language models.
The design-based contrasts were defined after descriptive evaluation rather than preregistered, and their overlapping components are not independent replications.
The results have also not yet been tested in an independent sample of respondents.
Bootstrap intervals reflect respondent resampling but not variation across repeated model runs, although the single-draw analysis indicates how much that variation moves the results (Supplementary Table~S19).
Finally, the follow-up decomposition separated bundles of features.
Layout included field labels and income formatting, and wording included grammatical perspective, which we tested separately under only one formal wording.
The decomposition identifies associations with these prompt components rather than internal model mechanisms.
The income check also shows that the size of the distributional distances depends on details of how the profile is presented, so these distances should be read as specific to our prompts, although the qualitative patterns, including the near absence of neutral predictions and each model's direction of bias, did not change (Supplementary Note~9).

For researchers using silicon samples, these findings support validating individual-level agreement and distributional fidelity separately.
Checks that need little or no human data can screen out unsuitable approaches early \cite{neumannShouldYouUse2026}, but whether a prompt reproduces individual answers and their distribution can only be established against real respondents.
A pilot sample from the target population can provide that reference, both for comparing prompts and for gauging the size and direction of their effects.
Designs that combine model predictions with a smaller sample of human respondents, who remain the reference, pursue a related goal \cite{broskaMixedSubjectsDesign2025}.
Responses used to choose or calibrate a prompt should still be kept apart from those used to evaluate the resulting procedure, and small pilots may leave rare responses and subgroups poorly assessed.
Our retrospective design does not establish performance when predictions are fixed before a new survey is fielded, and prospective validation remains an important test of the intended use.
Silicon samples are best treated as a complement to human data, and how adequate they are depends on the research task.
Because weighted $\kappa$ rewards dispersion as well as accuracy, changes in it are best reported together with a measure of error and the spread of the predictions.
A prompt that raises weighted agreement does not, by itself, validate the simulated distribution.

%% file: sections/methods.tex
\section*{Methods}
\subsection*{Data}
We used the 2022 General Social Survey (GSS), which was fielded during the COVID-19 pandemic and is based on a full-probability, multistage cluster sample of U.S. housing units \cite{davern2022GSSCrosssection2024}.
The 2022 respondents were extracted from the 1972--2024 cumulative data file (Release 2) through the GSS Data Explorer on 11 February 2026 \cite{davernGeneralSocialSurvey2025}.
We analysed two target items.
VAXSAFE asks respondents how far they agree or disagree with the statement ``Vaccines are safe'' on a five-point scale (1 = strongly agree, 5 = strongly disagree).
CONFEDVACY asks how much confidence respondents have in ``the federal government ensuring the safety of vaccines to protect the public against COVID-19'' on a three-point scale (1 = a great deal of confidence, 2 = only some confidence, 3 = hardly any confidence at all).
Both items were fielded in the GSS Next follow-on survey.

Respondent profiles were built from six sociodemographic variables, namely sex, age, years of education, inflation-adjusted family income (CONINC), religious preference and party identification.
CONINC is a recode of the bracketed family income question, expressed in constant 2000 dollars \cite{norcQuickGuideVariables2022}.
Consistent with this, each of its values below the top category in our sample equals the midpoint of a 2022 questionnaire income bracket multiplied by a common factor (Supplementary Note~9).
The prompts presented this value without stating its base year.
In the 2022 GSS, SEX was recorded by interviewer observation for face-to-face and telephone respondents and was derived from sex at birth and current gender identity for web respondents \cite{davern2022GSSCrosssection2024}.
Party identification was restricted to the seven-point Democrat--Republican scale, from 0 (strong Democrat) to 6 (strong Republican).

Of the 3{,}544 respondents in the 2022 GSS, 1{,}232 gave a valid response to VAXSAFE and 1{,}227 to CONFEDVACY (1{,}224 to both).
We retained respondents with valid responses to all five items of the 2022 vaccine module (CONPHARVACY, CONFEDVACY, VAXDOHARM, IMMUNBETR, VAXSAFE).
Two of these items (VAXDOHARM and IMMUNBETR) were administered only on Ballot C, so this restricted the sample to Ballot C respondents ($n = 399$).
Since 1988, GSS ballots have been administered to random subsamples within each survey year (\cite{smithGeneralSocialSurveys2019}, Appendix~Q).
The other three items were fielded in the GSS Next follow-on, so the sample was also limited to follow-on participants.
GSS Next was a web-only questionnaire offered to respondents who had completed the baseline interview before 30 November 2022.
Respondents from the AmeriSpeak oversample were not eligible.
The unweighted retention rate of GSS Next was 38.0\%, or 1{,}235 of 3{,}247 eligible baseline respondents (\cite{davern2022GSSCrosssection2024}, Table~2).
We then excluded respondents who identified with another party ($n = 20$) or gave no answer on party identification ($n = 2$), leaving $n = 377$.
Excluding respondents with missing age, sex or education left $n = 372$, excluding those with missing family income left $n = 341$, and removing one respondent with missing religious preference gave an analytic sample of $N = 340$ (Supplementary Table~S5).
The response distributions in the analytic sample closely resembled those of all valid 2022 responses.
For VAXSAFE, the category shares were 35.9/34.4/25.6/3.2/0.9\% in the analytic sample and 33.0/34.9/26.9/3.5/1.6\% among all valid responses.
For CONFEDVACY they were 42.6/37.6/19.7\% and 39.8/39.1/21.1\% (all unweighted).

We did not apply survey weights, including the GSS Next weights recommended for items from the follow-on survey.
Our aim is to evaluate individual predictions and the distribution of responses within this analytic sample, not to estimate population quantities.
All observed distributions in the bias analyses therefore refer to the unweighted analytic sample.

For analysis, religious preference was grouped as Protestant (including respondents identifying as ``Christian''), Catholic, None and Other (Jewish, Buddhism, Hinduism, Islam and other religions).
Jewish respondents ($n = 10$) were merged into Other because the cell was too small for stable estimation.
No respondent in the analytic sample reported another Eastern religion, Orthodox Christianity, a Native American religion or an inter- or nondenominational affiliation.
The grouping gave 135 Protestant, 76 Catholic, 97 None and 32 Other respondents.
It was applied only at the analysis stage, and the prompts presented each respondent's original religious category.

\subsection*{Language models}
We queried three models, GPT-5.4-mini (OpenAI), Claude Sonnet 4.6 (Anthropic) and Gemini 3 Flash (Google), through the API model identifiers \texttt{gpt-5.4-mini}, \texttt{claude-sonnet-4-6} and \texttt{gemini-3-flash-preview}.
These identifiers can point to updated model versions over time, so we report the dates of access.
In the main experiment, GPT-5.4-mini was accessed on 27--28 May 2026 for the base-prompt runs and on 5--6 August 2026 for the prompt-variant runs, Claude Sonnet 4.6 on 6 September 2026 and Gemini 3 Flash on 8--9 September 2026.
The follow-up prompts described below were run for all three models on 5 October 2026, and the income check, also described below, on 16--17 October 2026.
All dates are local dates in Taiwan (UTC+8).
Supplementary Table~S6 lists the model identifier and access date for every batch.
Because the base-prompt and prompt-variant runs of GPT-5.4-mini were collected about ten weeks apart, all prompt comparisons in the main experiment use the prompt-variant runs only, with P1 from the same period as the reference.
Within-batch follow-up contrasts used contemporaneous prompt runs, and contrasts F12 and F13 compared the October reruns with the corresponding original runs (Supplementary Table~S12).

All models were accessed through their official APIs without study-specific fine-tuning.
API calls, response parsing and aggregation were automated, and the authors designed the analyses and reviewed the resulting outputs.
Each query was a separate API call containing a single user message, with no system prompt and no conversation history, so no information carried over between respondents or runs.
All models were queried at temperature 0.2.
The maximum output length was set to 16 tokens for Claude Sonnet 4.6 and left at the provider default for the other two models.
No random seed was set.
Gemini 3 Flash was queried with \texttt{thinking\_budget=0}, and no additional reasoning configuration was supplied for GPT-5.4-mini or Claude Sonnet 4.6.
This setting describes the request, not the model's behaviour.
Current documentation for Gemini 3 models describes reasoning levels rather than a token budget and lists no setting that turns reasoning off, so we cannot confirm that no reasoning took place.
In the income-check batch, whose usage records included token counts, none of the 10{,}200 requests to each model reported reasoning tokens.
In the main-experiment notebooks examined, API exceptions were retried for Claude Sonnet 4.6 and Gemini 3 Flash, and empty replies were also retried for Gemini 3 Flash.
In the follow-up batch, API exceptions and empty replies were retried for all three models, with waiting between attempts.
Rate-limit errors from Gemini 3 Flash were handled separately, by waiting or by switching API keys, and interrupted batches could be resumed from saved records without repeating completed calls.
Reporting follows the GUIDE-LLM checklist for behavioural research with large language models \cite{feuerriegelReportingChecklistLarge2026}, and the completed checklist crosswalk is provided in Supplementary Table~S14.

\subsection*{Prompts}
Each prompt consisted of a profile, the survey item with its response options, and an instruction to answer with a single number.
The base prompt (P1) addressed the model in the second person.
It asked the model to assume that it was a US resident being interviewed in 2022 about public health during the COVID-19 pandemic, wrote the profile as a short narrative (for example, ``You are a 45 years old male, with a highest year of education being 4 years of college.
\ldots'') and asked for one number reflecting ``your attitude''.
The formal variants presented the same information as a record from a 2022 public health survey.
The profile was introduced as having been provided by the respondent and was listed as labelled fields (age, sex, educational attainment, inflation-adjusted annual family income, religious affiliation and political identification).
The item was introduced as a question put to respondents, and the instruction asked for the numeric response corresponding to the respondent's attitude.
Compared with the casual prompts, the formal variants therefore changed the register of the surrounding text, the grammatical perspective (the model answering as the respondent or reporting the respondent's answer) and the layout of the profile at the same time.
By register we mean the style of a survey instrument as opposed to conversational description, not politeness towards the model.
We call these variants formal survey-style prompts and do not attribute their effects to register alone.
The response options were identical in all prompts, but the item wording was not identical to the GSS questionnaire or across prompt styles.
All prompts asked whether the respondent agreed or disagreed ``that vaccine is safe'' rather than presenting the GSS statement ``Vaccines are safe''.
The formal prompt used in the main analysis introduced CONFEDVACY as a statement to agree with, although the item asks about confidence.
A follow-up variant corrected this introduction (P1FC, Supplementary Note~1).

Two factors were varied in a $3 \times 2$ design, the position of party identification within the profile (last, first or middle) and the prompt style (casual or formal).
The casual prompt with party identification last (P1) served as the base prompt.
The other variants are labelled P2 (party first) and P3 (party in the middle), with the suffix F marking the formal style.
The full text of all prompts is given in Supplementary Note~6.

\textit{Follow-up prompts.}
After the main experiment, we ran a second batch to examine the formal survey-style prompt, with party identification placed last throughout.
First, to separate profile layout from wording, we crossed the two in a $2 \times 2$ design.
P1S combined the casual, second-person wording of P1 with the structured profile of P1F, copied verbatim.
P1FN combined the formal wording of P1F with the narrative profile of P1, rewritten in the third person.
Together with P1 and P1F, both re-run in the same batch, these formed the four cells.
In this design, grammatical perspective moved with wording, and the field labels and the presentation of income moved with layout.
The structured profile labelled income as inflation-adjusted and gave it without a dollar sign, whereas the narrative profile gave it as a dollar amount without that label.
Second, to test whether the formal effects depended on the particular phrasing, we wrote two further formal versions, P1F2 and P1F3.
They kept the third-person perspective, the structured layout, the order of profile fields and the survey items, but changed the introduction, the field labels, the item introduction and the answer instruction.
A third version, P1F4, was identical to P1F3 except that it addressed the model in the second person, which also changed the tense of the opening clause (``was conducted'' became ``is being conducted'').
Before these prompts were run, we specified that the formal effect would be regarded as not tied to one wording if the changes in $\kappa$ and in the Wasserstein distance relative to P1 had the same sign as those of P1F in at least four of the six model--item combinations.
Finally, the formal prompts of the main experiment introduced the CONFEDVACY item as a request to ``indicate their level of agreement with the following statement'', although the item asks about confidence.
P1FC was identical to P1F except that this sentence was replaced by ``Respondents are asked the following question:'', and it was run for CONFEDVACY only.
P1F2 and P1F3 also introduced the question without the agreement framing.
Each follow-up prompt was run 3 times per respondent and item, and the full texts are given in Supplementary Note~6.

\textit{Income check.}
To examine whether the way income was presented affected the predictions, we ran a further batch with five prompt versions, two casual and three formal.
Within each style the versions differed only in the income line (Supplementary Note~9 and Supplementary Table~S15).
Each CONINC value in our sample corresponds to a single 2022 questionnaire bracket, so income could be presented either as the CONINC value or in the bracket wording of the questionnaire.
The batch included the base and formal prompts (P1 and P1F).
Two versions replaced the CONINC value with the nominal bracket, one casual (P122) and one formal (P1F22), and P1F22 labelled the bracket as 2021 dollars because the question refers to the previous year's income.
A fifth, formal version kept the CONINC value but stated its base year (P1F00).
Each version was run 3 times per respondent for both items and all three models.
The five versions were interleaved within each respondent so that any drift during the batch affected them equally, and all calls were completed within about two hours.
Comparisons were made only within this batch.

\subsection*{Prediction and aggregation}
Each respondent was queried individually.
The base prompt was run 10 times per respondent and item, and each of the six prompt variants 3 times.
The three P1 runs among the prompt variants were separate calls from the ten base runs, so P1 was queried 13 times per respondent and item in the main experiment.
Figs.~1--3 and Table~1 use the ten base runs, and the prompt comparisons of the main experiment use the three P1 runs as the reference.
The follow-up batch and the income check each re-ran their own reference prompts, and every within-batch contrast uses the reference from the same batch.
Only contrasts F12 and F13 in Supplementary Table~S12 deliberately compare the October reruns with the original variant runs.
For each model, respondent and prompt, the prediction was the mode across runs.
Ties went to the lower value, that is, the more agreeing or more confident response option, and a three-way tie in the three-run variants went to the lowest of the three values.
They were rare, occurring for at most 4.4\% of respondents in any model--item--prompt condition and for fewer than 0.3\% in the three-run prompt variants.
Resolving ties towards the higher value instead changed weighted $\kappa$ by at most 0.01 and did not alter the best-performing prompt for any model or item.
Responses were parsed as integers, and any response outside the valid scale range was to be treated as missing.
The final datasets contained no parsing failures and no responses outside the valid range, and a modal prediction was obtained for every respondent in every condition in the main, follow-up and income-check batches.
Each run's parsed response was kept separately and consolidated into a master dataset, together with the modal predictions used in the analyses.

\textit{Sensitivity to aggregation.}
We repeated the main analyses with two alternatives to the mode.
In the first, one run was drawn at random for each respondent, and this was repeated 500 times.
We report the mean across draws and the 2.5th and 97.5th percentiles, which reflect variation across draws only.
In the second, the response frequencies across runs were averaged over respondents to give each model's predicted distribution, and only the Wasserstein distance was computed, since this procedure gives no prediction for any individual.
Its prompt contrasts were assessed with the same paired bootstrap as the main analysis.
Because each prompt variant had only three runs per respondent, the single-draw analysis is coarse for the variants.

\subsection*{Statistical benchmark}
The prompts conveyed only the six sociodemographic variables, so a regression model using the same variables indicates how much individual-level information those variables carry about each item \cite{huQuantifyingPersonaEffect2024}.
We therefore fitted an ordered logit model \cite{mccullaghRegressionModelsOrdinal1980} for each item on the full analytic sample, with religion dummy-coded and None as the reference category.
Each respondent was assigned the category with the highest predicted probability.
Models were estimated with \texttt{statsmodels} (version 0.13.2) \cite{seaboldStatsmodelsEconometricStatistical2010}.
To avoid evaluating the benchmark on the respondents used to fit it, we obtained out-of-fold predictions by tenfold cross-validation, stratified by the observed response and repeated 20 times with different random splits.
Each respondent's benchmark prediction was the mode of the 20 out-of-fold predictions, with ties resolved towards the lower category as for the language models.
Across repetitions, $\kappa$ ranged from 0.31 to 0.37 for VAXSAFE and from 0.45 to 0.50 for CONFEDVACY.
Fitting and evaluating the model on the full sample gave $\kappa$ of 0.39 and 0.51.

\subsection*{Evaluation metrics}
\textit{Reliability.}
Consistency across the 10 base-prompt runs was assessed with Krippendorff's $\alpha$ for ordinal data \cite{krippendorffContentAnalysisIntroduction2019} and with ICC(2,1), the two-way random-effects, absolute-agreement, single-rater intraclass correlation \cite{shroutIntraclassCorrelationsUses1979,mcgrawFormingInferencesIntraclass1996} (Supplementary Table~S2).

\textit{Validity.}
Agreement between predicted and observed responses was measured with quadratic-weighted Cohen's $\kappa$ \cite{cohenWeightedKappaNominal1968} and mean absolute error (MAE).
Our implementation of weighted $\kappa$ was checked against scikit-learn \cite{JMLR:v12:pedregosa11a}.
Spearman's rank correlation largely overlaps with weighted $\kappa$ and is reported in Supplementary Table~S3.

\textit{Decomposition of weighted agreement.}
With quadratic weights, $\kappa = 1 - \mathrm{MSE}/D$, where MSE is the mean squared difference between predicted and observed responses and $D = \mathrm{Var}(Y) + \mathrm{Var}(P) + (\bar{Y} - \bar{P})^2$ is the value MSE would take if predictions were unrelated to the observed responses.
Here $Y$ and $P$ are the observed and predicted responses, and both variances use the divisor $N$.
A prompt change can therefore raise $\kappa$ by lowering MSE or by raising $D$, for example by spreading the predictions more widely.
For each prompt contrast, we split the change in $\kappa$ into a part due to the change in MSE and a part due to the change in $D$, averaging over the two orders in which the two changes can be applied, so that the parts sum exactly to the change in $\kappa$ (Supplementary Table~S18).

\textit{Distributional bias.}
Predicted and observed response distributions were compared using the Wasserstein-1 distance, the directional shift (mean predicted minus mean observed response), and boundary avoidance (the proportion of observed minus predicted responses in the two extreme categories).

\textit{Error structure.}
To examine which respondent characteristics were associated with prediction error, we regressed the absolute prediction error under the base prompt on the six sociodemographic variables by ordinary least squares, separately for each model and item.
Age, years of education, family income and party identification entered as continuous variables.
Sex entered as an indicator for female respondents, and religion as indicators with no religion as the reference category.
Standard errors were heteroskedasticity-consistent (HC3) \cite{mackinnonHeteroskedasticityconsistentCovarianceMatrix1985}, because the variance of absolute errors is likely to differ across respondents.

\textit{Subgroup analysis.}
MAE under the base prompt was computed within subgroups defined by age (18--35, 36--55, 56+), sex, years of education ($\le 11$, 12--15, $\ge 16$), income tertiles, party identification (Democrat 0--1, Independent 2--4, Republican 5--6) and religious group.

\subsection*{Uncertainty and comparisons}
All metrics were computed on the full analytic sample.
Their 95\% confidence intervals came from a percentile bootstrap \cite{efronIntroductionBootstrap1993} with 10{,}000 resamples of respondents and a fixed random seed of 42.
Differences between two predictors evaluated on the same respondents were estimated with a paired bootstrap, in which both predictors were scored on the same resampled respondents.
We regarded a difference as distinguishable from zero when the 95\% interval of the paired difference excluded zero, and we did not infer differences from the overlap of separate intervals.
A difference survived correction when its Benjamini--Hochberg $q$ was below 0.05 within its family of comparisons.
Conclusions stated in the abstract, section headings and Discussion rest on the corrected results.
Differences between models in a prompt effect were estimated in the same way.
On each resample, both models' changes were computed on the same respondents and their difference was taken.
For each metric, the 24 comparisons formed by two items, four contrasts and three pairs of models were treated as one family for multiplicity adjustment.
The benchmark's cross-validated predictions were held fixed during resampling rather than recomputed, so these intervals do not include uncertainty from re-estimating the regression coefficients.

Each model was compared with the benchmark under the base prompt and, descriptively, under the prompt variant with the highest full-sample $\kappa$ for that model and item.
This variant was selected on the same data, so the second comparison favours the model.
Complete comparisons are reported in Supplementary Table~S10.

Prompt effects were assessed with four design-based contrasts corresponding to the two design factors, each estimated within model and item on the prompt-variant runs.
Two contrasts concern the formal survey-style prompt.
One is P1F minus P1, which changes the prompt style while holding party position at its base value.
The other is the mean of the three formal variants minus the mean of the three casual variants.
A third contrast concerns position (P2 minus P1), and the fourth combines both changes (P2F minus P1).
The contrasts were defined after the prompt variants had been evaluated descriptively, so we treat them as a structured summary of the design rather than as preregistered tests.
Four contrasts across three models, two items and several metrics yield many intervals, so we interpret them as a pattern across models and items.
To show which individual differences hold up under multiple testing, we also report false discovery rate (Benjamini--Hochberg \cite{benjaminiControllingFalseDiscovery1995}) and Bonferroni adjustments for each metric.
The adjustments were computed within the family of 24 contrasts formed by four contrasts, three models and two items, from two-sided bootstrap $p$ values (Supplementary Table~S9).
Each $p$ value was twice the smaller of the proportions of resampled differences at or below zero and at or above zero, with one added to the numerator and to the denominator of each proportion, and was capped at 1.
With 10{,}000 resamples the smallest attainable $p$ value is about 0.0002.
Because the $p$ values and the percentile intervals are computed differently, an interval that only just excludes zero can accompany a $p$ value slightly above 0.05.

In the follow-up batch, the $2 \times 2$ design was summarized by the main effect of the structured layout (the mean of P1S minus P1 and P1F minus P1FN), the main effect of the formal wording (the mean of P1FN minus P1 and P1F minus P1S) and their interaction.
The two main effects sum to P1F minus P1.
Several further comparisons were estimated in the same way.
They comprised the wording comparisons (P1F2 and P1F3 against P1 and against P1F), the perspective comparison (P1F3 minus P1F4), the corrected item introduction (P1FC minus P1F) and the change in each re-run prompt between batches.
False discovery rate adjustment was applied within each metric across the 87 contrasts of this batch (Supplementary Table~S12).
For each model these comprised 13 contrasts for VAXSAFE, namely the full rewrite (P1F minus P1), the four wording comparisons, the three effects of the $2 \times 2$ design, three steps of a decomposition from P1 through P1S and P1F4 to P1F3, and the two between-batch changes, and 16 for CONFEDVACY, which added three comparisons involving P1FC.
These analyses were conducted after the main results were known and are reported as a follow-up decomposition.
In the income check, four comparisons were estimated in the same way (P122 minus P1, P1F00 minus P1F, P1F22 minus P1F, and P1F22 minus P1F00).
False discovery rate adjustment was applied within each metric across the 24 comparisons formed by four comparisons, three models and two items (Supplementary Table~S17).
All bootstrap intervals reflect sampling variation across respondents only and do not include variation across repeated model runs.

\subsection*{Recall probe}
As a preliminary probe of whether the models could recall the survey results, we asked each model, without any respondent profile, to report the published GSS 2022 response distribution for each item, and told it to answer only if it genuinely recalled the figures.
Each model was queried 10 times per item under the same settings as the main analysis.
In every run, all three models said that they did not recall the distribution.
This check addresses recall of aggregate distributions only.
It does not rule out exposure to respondent-level GSS data during training.

\subsection*{Ethics}
This study is a secondary analysis of publicly available, de-identified data from the General Social Survey.
It involved no interaction with human participants and no access to identifiable private information.
Institutional review board review was therefore not sought.

\subsection*{Use of AI tools in manuscript preparation}
Claude (Anthropic; Opus 5.5 and Fable 5.2), ChatGPT (OpenAI; GPT-6 Astra and GPT-6.2 Astra) and Gemini (Google; Gemini 4 Argon and Gemini 4 Neon) assisted with manuscript preparation.
Claude was used to draft and revise text from the authors' notes and analysis outputs, to assist with writing and reviewing analysis code, including modularization and package-specific implementation, and code used to check the reported numbers against the analysis outputs, and to check cited claims against the full texts of the cited articles.
ChatGPT was used to revise the manuscript, check the correspondence between claims and references, assist with analysis and export code, cross-check reported descriptive statistics against author-provided data and review the full manuscript.
Gemini was used only to review the full manuscript.
The authors reviewed and edited all generated content and take full responsibility for the content of this publication.
This use is separate from the three language models evaluated in this study, which were accessed only through their APIs as described above.

\section*{Data availability}
The 2022 GSS data are publicly available from NORC at the University of Chicago (\url{https://gss.norc.org}).
Parsed model responses for all respondents, prompts and runs, together with the modal predictions used in the analyses, will be made publicly available upon publication.
Model response text was additionally retained for some batches, but not for all runs.
Complete API response payloads were not retained.

\section*{Code availability}
Code for querying the models, aggregating predictions, and reproducing all analyses and figures will be made publicly available upon publication.